\documentclass[journal]{IEEEtran}

\usepackage{cite}
\usepackage{balance}

\ifCLASSINFOpdf
  \usepackage[pdftex]{graphicx}
  \graphicspath{{../pdf/}{../png/}}
  \DeclareGraphicsExtensions{.pdf,.jpeg,.png}
\else
\fi
\usepackage{framed}

\usepackage{amsmath}
\usepackage{amsfonts}
\DeclareMathOperator{\mean}{mean}

\DeclareMathOperator{\csch}{csch}
\DeclareMathOperator{\sech}{sech}
\DeclareMathOperator{\polylog}{Li}
\DeclareMathOperator{\relu}{relu}

\usepackage{algorithmic}
\usepackage{algorithm}
\usepackage{array}
\usepackage{tabularx}
\usepackage{afterpage}
\def\arraystretch{1.5}
\usepackage{makecell}

\usepackage{url}
\begin{document}
%
% paper title
% Titles are generally capitalized except for words such as a, an, and, as,
% at, but, by, for, in, nor, of, on, or, the, to and up, which are usually
% not capitalized unless they are the first or last word of the title.
% Linebreaks \\ can be used within to get better formatting as desired.
% Do not put math or special symbols in the title.
\title{AMITE: A Novel Polynomial Expansion for
Analyzing Neural Network Nonlinearities}
%
%
% author names and IEEE memberships
% note positions of commas and nonbreaking spaces ( ~ ) LaTeX will not break
% a structure at a ~ so this keeps an author's name from being broken across
% two lines.
% use \thanks{} to gain access to the first footnote area
% a separate \thanks must be used for each paragraph as LaTeX2e's \thanks
% was not built to handle multiple paragraphs
%

\author{Mauro~J.~Sanchirico~III,~
        Xun~Jiao,~and~C. Nataraj% <-this % stops a space
\thanks{M. Sanchirico is with the Lockheed Martin Artificial Intelligence Center, Mt Laurel Township, NJ 08054 USA and the Department of Electrical and Computer Engineering, Villanova University, Villanova PA, 19085 USA}
\thanks{X. Jiao is with the Department
of Electrical and Computer Engineering, Villanova University, Villanova, PA, 19085 USA}% <-this % stops a space
\thanks{C. Nataraj is with the Villanova Center for Analytics of Dynamic Systems (VCADS), Villanova University, Villanova, PA, 19085 USA}% <-this % stops a space
}

% make the title area
\maketitle

% As a general rule, do not put math, special symbols or citations
% in the abstract or keywords.
\begin{abstract}
Polynomial expansions are important in the analysis of neural network nonlinearities. They have been applied thereto addressing well-known difficulties in verification, explainability, and security. Existing approaches span classical Taylor and Chebyshev methods, asymptotics, and many numerical approaches.  We find that while these individually have useful properties such as exact error formulas, adjustable domain, and robustness to undefined derivatives, there are no approaches that provide a consistent method yielding an expansion with all these properties.  To address this, we develop an analytically modified integral transform expansion (AMITE), a novel expansion via integral transforms modified using derived criteria for convergence.  We show the general expansion and then demonstrate application for two popular activation functions, hyperbolic tangent and rectified linear units. Compared with existing expansions (i.e., Chebyshev, Taylor, and numerical) employed to this end, AMITE is the first to provide six previously mutually exclusive desired expansion properties such as exact formulas for the coefficients and exact expansion errors (Table \ref{table_qualitative_comparison}).  We demonstrate the effectiveness of AMITE in two case studies.  First, a multivariate polynomial form is efficiently extracted from a single hidden layer black-box Multi-Layer Perceptron (MLP) to facilitate equivalence testing from noisy stimulus-response pairs.  Second, a variety of Feed-Forward Neural Network (FFNN) architectures having between 3 and 7 layers are range bounded using Taylor models improved by the AMITE polynomials and error formulas.  AMITE presents a new dimension of expansion methods suitable for analysis/approximation of nonlinearities in neural networks, opening new directions and opportunities for the theoretical analysis and systematic testing of neural networks.
\end{abstract}

% Note that keywords are not normally used for peerreview papers.
\begin{IEEEkeywords}
Neural Networks, Equivalence, Taylor, Fourier, Polynomial, Approximation.
\end{IEEEkeywords}

% For peer review papers, you can put extra information on the cover
% page as needed:
% \ifCLASSOPTIONpeerreview
% \begin{center} \bfseries EDICS Category: 3-BBND \end{center}
% \fi
%
% For peerreview papers, this IEEEtran command inserts a page break and
% creates the second title. It will be ignored for other modes.
\IEEEpeerreviewmaketitle

\section{Introduction}
\label{section_introduction}
% The very first letter is a 2 line initial drop letter followed
% by the rest of the first word in caps.
% 
% form to use if the first word consists of a single letter:
% \IEEEPARstart{A}{demo} file is ....
% 
% form to use if you need the single drop letter followed by
% normal text (unknown if ever used by the IEEE):
% \IEEEPARstart{A}{}demo file is ....
% 
% Some journals put the first two words in caps:
% \IEEEPARstart{T}{his demo} file is ....
% 
% Here we have the typical use of a "T" for an initial drop letter
% and "HIS" in caps to complete the first word.
\IEEEPARstart{I}{n} the design and implementation of
nonlinear systems, polynomial expansions 
provide an important means of decomposing nonlinear
functions into a form facilitating analysis. Both classical and modern neural networks have inherent nonlinearities in their activation functions such as hyperbolic tangent and rectified linear units. 
Various polynomial expansions and approximations have been
applied to neural network activations enabling explainable, verifiable, and secure deployment of neural systems, with \cite{montavon_explaining_2017}, \cite{dutta_reachability_2019},
and \cite{obla_effective_2020} being three recent examples
respectively.  The motivation to develop explainability 
and verification techniques is further enhanced by proposed or
studied use of neural networks for critical systems such as unmanned
aerial vehicles \cite{lai_adaptive_2016,nodland_neural_2013},
hypersonic vehicles \cite{xu_dob-based_2017}, or even
air-to-air combat in recent work \cite{pope_hierarchical_2021}.
Meanwhile, use of neural networks in healthcare applications, such as
\cite{shen_deep_2019}, further drives the need for
secure inference techniques to protect third party private
information \cite{kwabena_mscryptonet:_2019,obla_effective_2020}.
The polynomial expansion of neural network activation functions
is a method employed in addressing all of these diverse and
important challenges.

\begin{table}
    \caption{
        Qualitative Properties Considered for a Polynomial
        Expansion of the Form:
        $\varphi(v) = \sum_m \alpha_m^M v^m + H_M(v)$
    }
    \begin{tabularx}{0.98\columnwidth}{
      >{\hsize=.4\hsize}X
      >{\hsize=1.4\hsize}X
    }
        \hline
        \textit{Property} & \textit{Description} \\
        \hline
        \hline
        \textit{Exact \mbox{Coeffs.}} 
        &
            Precise analytic equations available
            for the coefficients, $\alpha_m$, for the nonlinearities
            considered.
        \\
        \hline
        \textit{Explicit Exact Errors} 
        &
            Precise analytic equations are explicitly worked for
            the expansion errors, $H_M(v)$.
        \\
        \hline
        \textit{Monic Form}
        &
            The expansion is formulated directly as a sum of
            monomials, i.e., $\sum_m \alpha_m v^m$ and not as
            a sum of other polynomials or piecewise polynomials.
        \\
        \hline
        \textit{Adjustable Domain}
        &
            The approximation $\varphi(v) = \sum_m^M \alpha_m v^m$
            is valid for any desired domain of validity $|v| < V$
            without transforming or re-scaling $v$.
        \\
        \hline
        \textit{Handles undef. derivs.}
        &
            The expansion is valid if some of the
            derivatives of $\varphi$ are undefined.
        \\
        \hline
        \textit{Consistent Method}
        &
            The method can be applied consistently to a wide
            variety of functions without ad-hoc modifications.
        \\
        \hline
    \end{tabularx}
    \label{table_expansion_properties_considered}
\end{table}

\begin{table*}[t]
% increase table row spacing, adjust to taste
\renewcommand{\arraystretch}{1.3}
% if using array.sty, it might be a good idea to tweak the value of
% \extrarowheight as needed to properly center the text within the cells
\caption{Qualitative Comparison of Polynomial Expansion Techniques
Available for Neural Network Nonlinearities}
\label{table_qualitative_comparison}
\centering
% Some packages, such as MDW tools, offer better commands for making tables
% than the plain LaTeX2e tabular which is used here.
\begin{tabularx}{0.98\textwidth}{
    |c|
    | >{\centering\arraybackslash}X
    | >{\centering\arraybackslash}X
    | >{\centering\arraybackslash}X
    | >{\centering\arraybackslash}X
    | >{\centering\arraybackslash}X
    | >{\centering\arraybackslash}X
    | >{\centering\arraybackslash}X
    | >{\centering\arraybackslash}X
    | }
\hline
\ & \textit{Exact Coeffs.}
  & \textit{Explicit Exact Errors}
  & \textit{Monic Form}
  & \textit{Adjustable Domain}
  & \textit{Handles Undef. Derivs.}
  & \textit{Consistent Method} \\
\hline
\hline
AMITE                                                    & Y & Y & Y & Y & Y & Y \\
\hline
\hline
Asymptotics \cite{noauthor_hyperbolic_nodate}           & Y & \ & \ & \ & Y & \ \\
\hline
Bernstein \cite{huang_reachnn:_2019}                    & Y & \ & \ & Y & Y & Y \\
\hline
Chebyshev \cite{vlcek_chebyshev_2012}                   & Y & \ & \ & \ & Y & Y \\
\hline
Exponential Chebyshev \cite{koc_new_2013}               & Y & \ & \ & Y & Y & Y \\
\hline
Fast Fourier-Legendre \cite{alpert_fast_1991}           & \ & \ & \ & \ & Y & Y \\ 
\hline
Least Squares Regression \cite{obla_effective_2020}     & \ & \ & Y & Y & Y & Y \\
\hline
Minimax w. Chebyshev \cite{obla_effective_2020, lee_near-optimal_2020} 
                                                        & \ & \ & \ & Y & Y & Y \\
\hline
Piecewise \cite{baptista_low-resource_2013}             & \ & \ & \ & Y & Y & \ \\
\hline
Recursive Regression \cite{dutta_reachability_2019}     & \ & \ & \ & Y & Y & Y \\
\hline
Remez / Modified Remez \cite{lee_high-precision_2020, lee_high-precision_2020-1}
                                                        & \ & \ & \ & Y & Y & Y \\
\hline
Taylor\cite{bateman_higher_1953}                        & Y & Y & Y & \ & \ & Y \\
\hline

\hline
\end{tabularx}
\end{table*}

%To frame our motivations, we briefly review a sample of the panoply of polynomial expansions suitable for approximation of neural nonlinearity. 
The existing polynomial expansions of neural nonlinearity can be categorized as follows. 
First, the well-known Taylor series is employed often
\cite{engelbrecht_using_2000,xiaoyun_new_neural_nets_2009,balduzzi_neural_2017,montavon_explaining_2017,guo_novel_2017,gaikwad_pruning_2018}.
While Taylor series provides the advantage of exact formulas
for coefficients and errors and often provides a useful local
approximation, its domain of validity is limited to $|v| < \frac{\pi}{2}$
for $\tanh(v)$ nonlinearities \cite{bateman_higher_1953} and it
requires that a function have defined derivatives in the expansion domain.
A well-known identity for large arguments is given:
$\tanh(v) = 1 + 2\sum_{m=1}^{\infty}(-1)^m e^{-2 m v}\ \forall\ v > 0$
\cite{gradshtein_table_2007}, and can be manipulated into a power series
like form by expanding $e^{-2 m v}$. However, the result
is still not valid for small $v$.  Similarly, many asymptotic series
are available for nonlinear functions such as $\tanh$ but
often have restricted domains (e.g., large arguments) 
\cite{noauthor_hyperbolic_nodate}, do not have a monic form,
and are not generally differentiable
\cite{gradshtein_table_2007}.

A Chebyshev approximation can be applied using recursive formulas
for the coefficients \cite{vlcek_chebyshev_2012}.  References
employing Chebyshev approximations or comparing them to other
approximations for neural nonlinearities include
\cite{aboulatta_stabilizing_2019, lee_near-optimal_2020} and
\cite{obla_effective_2020}.
Alternatively, the Fourier-Legendre series \cite{bojanic_rate_1981}
has well-known methods for fast coefficient computation
\cite{alpert_fast_1991} and provides a fast rate of convergence.
The Bernstein basis is employed by
Huang et al. in ReachNN to perform reachability analysis
for neural controlled systems \cite{huang_reachnn:_2019}.
Overall, the related polynomial basis techniques via Chebyshev,
Fourier-Legendre, Bernstein and similar methods provide accurate
approximation with the useful property that formulas for coefficients
can be exact using some methods.  However, while most methods employing
such techniques provide upper bounds on the error, exact error formulas
have not yet been provided for the expansions of the neural nonlinearities
considered.

If analytic formulas are not required, many numerical techniques
are available to approximate a nonlinearity.  Obla et al. compare
least squares and best uniform approximation
(i.e., minimax approximation) against Taylor series
\cite{obla_effective_2020}. Lee et al. also use a minimax
approximation with a Chebyshev basis \cite{lee_near-optimal_2020},
improving upon results from a modified Remez algorithm
\cite{lee_high-precision_2020}.
Numerically fit piecewise polynomials are suited for efficient
implementation of a nonlinear function such as
$\tanh$ over a wide domain
\cite{namin_efficient_2009,armato_low-error_2011}.
Piecewise approximations can be combined with other methods,
e.g., Chebyshev \cite{baptista_low-resource_2013}.

Piecewise linear polynomials
specifically are well suited to $\relu$ nonlinearities which are a
special case thereof. They are key components of the Marabou
framework for verification of deep neural networks
\cite{katz_marabou_2019} and employed by Dutta et al. to
perform reachability analysis via regressive polynomial rule inference
\cite{dutta_reachability_2019}.  Dutta et al. further
reduce the expansion error by repeatedly fitting a low order polynomial
and fitting a new polynomial to the residual. This approach keeps 
the resulting polynomial order low but does not lend itself to extraction of
exact error formulas, which the authors cite as a challenge and then
solve by further reduction to a piecewise linear model
\cite{dutta_reachability_2019}.

To address the limitations of many existing polynomial approximations,
we employ a special functions approach to develop a new technique,
applicable to a wide class of neural network 
nonlinearities and having many of the desirable properties of the
above expansions including many that were previously mutually
exclusive.  Motivated by applications noted above, we provide,
for the first time, the derivation of explicit
exact error formulas and useful analytic approximations thereof.
To facilitate mathematical manipulation, we provide a monic
polynomial form like that of the Taylor series and an adjustable
domain of convergence that does not require transformation
or re-scaling.  We provide exact coefficients that do not
require iteration or recursion to compute and avoid the main
disadvantage of the Taylor series by remaining robust to undefined
derivatives.  Furthermore, the method provides a consistent
approach which can be readily applied to a wide variety of
nonlinear functions without \emph{ad hoc} manipulation for a specific
nonlinearity type. Exploitation of these new properties is then demonstrated
in the case studies that follow.

The expansion is developed using integral transforms of the generalized
Fourier type, modified according to derived criteria for convergence and
as such is called the analytically modified integral transform expansion
(AMITE) technique. The qualitative expansion properties that we
consider are defined in Table \ref{table_expansion_properties_considered}.
We present a comprehensive comparison of AMITE with existing state-of-the-art
expansion methods with respect to these properties in Table
\ref{table_qualitative_comparison}.

We summarize our contributions in this paper as follows.
\begin{itemize}
    \item \textbf{Novel polynomial expansion and errors:}
    We develop a novel polynomial expansion technique,
the analytically modified integral transform expansion
(AMITE), via a special functions approach to enable
decomposing neural network activation functions
into polynomial forms having several advantageous properties,
listed in Table \ref{table_qualitative_comparison},
many of which would otherwise be mutually exclusive.
We show the general expansion method and its specific
employment for the hyperbolic tangent and ReLU.
Exact error formulas are derived and analyzed
via exploitation of a number of non-trivial relationships
amongst the special and elementary functions.

\item \textbf{Black-box MLP equivalence testing example:}
Using a single hidden layer MLP as a first case study, we
demonstrate use of the AMITE technique in
a black-box system identification
based equivalency test for checking that a given MLP is
output-equivalent to another from noisy stimulus-response pairs.
While similar methods would require piecewise polynomials
or numerical fits due to the limitations of existing
expansions available for neural nonlinearities, we show
that we can efficiently expand the entire MLP as a single
multivariate polynomial approximation. This is enabled by
the adjustable domain of convergence and simple monic form,
and is advantageous for facilitating exact error analysis.

\item \textbf{Improved range bounding for deep FFNNs:}
Using the problem of range bounding the output of a deep FFNN
over a set of input domain intervals as a second case study,
we construct an improved Taylor model representation for deep FFNNs
wherein the AMITE polynomials and exact error formulas are
specifically exploited to provide tighter output range bounds
over larger domain intervals than possible with a conventional Taylor model
(i.e., a Taylor model based on the Taylor series and its error bounds).
A method for speeding up computation of the range bounds using
the AMITE's analytic error approximations while retaining the rigor of
exact error handling is provided for the first
time.  These improvements combined with the newly developed
exact formulas imply that the AMITE technique
is a useful complement to the existing repertoire of
polynomial expansions employed in the analysis of
neural systems.

\end{itemize}

\textbf{Organization of the paper:}
Section \ref{section_definitions} defines the notation.
The main expansions
and errors are developed in Section
\ref{section_improved_expansions} and expansion results
are summarized in Section \ref{section_expansion_results}.
The case study for equivalence testing a single hidden layer MLP
is presented in Section \ref{section_network_replication} and the
case study for range bounding a deep FFNN is presented in Section \ref{section_range_bounding}.
Related work is reviewed in Section \ref{section_related_work}
and Section \ref{section_conclusion} concludes the paper.

\section{Definitions and Problem Formulation}
\label{section_definitions}
The hyperbolic tangent and ReLU are defined:
$\tanh(v) = \frac{e^v - e^{-v}}{e^v + e^{-v}}$,
$\relu(v) = \frac{|v| + v}{2}$.
When developing approximations of activation functions $\varphi$
we denote their coefficients $\alpha_m$
\begin{equation}
    \label{eq_alpha_expansion}
    \varphi(v) \simeq \sum_{m=0}^{M} \alpha_m v^m.
\end{equation}

We consider FFNNs having $N_I$ inputs, $N_L$ layers, and $N_H^{\{l\}}$ hidden
neurons per layer $l$. We use $\alpha$ to denote activation function
coefficients and $\beta$ to denote the collection of weights
and biases
$\beta = \left\{W^{\{0\}}, \dots, W^{\{N_L\}}, b^{\{0\}}, \dots, b^{\{N_L\}} \right\}$
such that each network layer has the input-output relationship:
\begin{equation}
    \label{eq_mlp_input_output}
    y\left(\alpha, \beta, x\right) = W^{\{l\}} \varphi\left(
        W^{\{l-1\}} x + b^{\{l-1\}}\right) + b^{\{l\}}
\end{equation}
Linear activations are used on the last layer.  The input to hidden neuron $n$ is denoted
\begin{equation}
    \label{eq_hidden_neuron_input}
    v_n = b_n^{\{l-1\}} + \sum_{i=1}^{N_I} w_{n,i}^{\{l-1\}}x_i \quad
    \forall \quad n \in [1, N_H^{\{l\}}].
\end{equation}
%Hidden layer weights and biases are denoted
%$W^{\{l\}} = \left[w_{n,i}^{\{l\}}\right] \in \mathcal{M}^{N_H \times N_I}$
%and $b^{\{0\}} = \left[b_{n}\right] \in \mathcal{M}^{N_H \times 1}$. Hidden
%layer weights and bias are denoted
%$w^{\{1\}} = \left[w_{n}^{\{1\}}\right] \in \mathcal{M}^{1 \times N_H}$
%and $b^{\{1\}} \in \mathcal{M}^{1 \times 1}$.  The network has the activation
%function $\varphi(v)$.  The weights and biases for a given network are
%collectively denoted
%$\beta = \left\{ W^{\{0\}}, w^{\{1\}}, b^{\{0\}}, b^{\{1\}} \right\}$.

The standard definition of a Taylor model, $\mathcal{T}$,
due to Berz \cite{berz_taylor_1997,berz_taylor_nodate} for a function,
$\varphi$, is employed, i.e.,
\begin{equation}
    \mathcal{T}_M^{\{\varphi\}}= \left\{
        \left[\alpha_m\right],
        (e_{\mathrm{min}}, e_{\mathrm{max}})\right\},
\end{equation}
where $\left[\alpha_m\right]$ and
$\left(e_{\mathrm{min}}, e_{\mathrm{min}}\right)$ denote the polynomial
coefficients of $\varphi$ and the rigorous error interval
of the approximation.  While Taylor series is
usually employed to derive $\alpha$, other polynomial approximation
methods (surveyed in \ref{section_introduction}) have been employed.
In the second case study, AMITE and Taylor polynomials are compared
for the Taylor modeling of deep FFNNs.
Standard rules \cite{berz_taylor_1997,berz_taylor_nodate}
for addition, multiplication, and composition of Taylor models are employed.
Interval arithmetic operations for handling errors are implemented as a
simplified subset of IEEE 1788.1 \cite{noauthor_ieee_2018}. We assume that nonlinearities, $\varphi$,
to be analyzed can be represented by integral transforms in the form of
$\varphi(v) = \int_{\mathbb{R}}\Phi(\xi)f(\xi v)\mathrm{d}\xi$
wherein the kernels have Taylor series convergent on $(-\infty, \infty)$
of the form $f(\xi v) = \sum_{m=0}^\infty \tilde{f}(m) (\xi v)^m$.

% needed in second column of first page if using \IEEEpubid
%\IEEEpubidadjcol
\section{AMITE Development}
\label{section_improved_expansions}
To achieve the expansion properties shown in Table
\ref{table_qualitative_comparison} for the AMITE technique
we seek expressions of the forms:
\begin{align}
    \tanh(v) &\simeq \sum_{m=0}^{M} T_{2m+1}(M) v^{2m+1} , 
        \quad &\forall \quad |v| < V , \label{eq_tanh_approx_form}\\
    \relu(v) &\simeq \frac{v}{2} + \sum_{m=0}^{M} R_{2m}(M) v^{2m} ,
        \quad &\forall \quad |v| < V . \label{eq_relu_approx_form}
\end{align}
In (\ref{eq_tanh_approx_form}) and (\ref{eq_relu_approx_form}) we
require explicit formulas for evaluation to arbitrary
precision, yielding expansions valid for a tunable desired
bound of validity $V$.
The coefficients are allowed to depend on the order of
expansion $M$.
We present the general expansion
method which can be applied to a wide class of
activations and then apply the
method explicitly to $\tanh(v)$, $\relu(v)$, and finally
to an entire network layer.

\subsection{General Expansion Technique}
\label{section_method_technique}
The general method for obtaining the coefficients
is as follows.  First, the function $\varphi$ to be
approximated is expressed via a general Fourier integral transform of the form:
$\varphi(v) = \int_\mathbb{R} \! \Phi(\xi) f(\xi v) \mathrm{d}\xi.$
The kernel $f(\xi v)$ is then expanded via the truncated Taylor series
\begin{align}
    f(\xi v) &\simeq \sum_{m=0}^M \tilde{f}(m) (\xi v)^m
        + \Upsilon_M^{\{f\}}(\xi v),
        \label{eq_taylor_f} \\
    \varphi(v) &\simeq \int_\mathbb{R} \!
        \Phi(\xi) \sum_{m=0}^M \tilde{f}(m) (\xi v)^m \mathrm{d}\xi.
        \label{eq_phi_v_integral}
\end{align}
The remainder $\Upsilon_M^{\{f\}}(\xi v)$ is factored into parts
$P_M(\xi v)$ and $Q_M(\xi v)$
that increase with and without bound respectively
\begin{equation}
\label{eq_remainder_factorization}
    \Upsilon_M^{\{f\}}(\xi v) = P_M(\xi v) Q_M(\xi v) .
\end{equation}

The dominating term $Q_M(\xi v)$ is set equal to a constant $C$ and
solved for $\xi$ to determine a suitable limit $\xi = \rho(M)$ of
integration for (\ref{eq_phi_v_integral}) which remains in the
region of convergence of (\ref{eq_taylor_f}), i.e., $\xi = Q_M^{-1}(C) / v$.
In cases where $v$ varies, the limit of
integration can be defined with respect to its maximum $V$:
$\rho_V(M) = Q_M^{-1}(C)/V$.
Then Eq. (\ref{eq_phi_v_integral}) can be rewritten
\begin{equation}
    \varphi(v) \simeq
        \sum_{m=0}^M \tilde{f}(m) v^m
            \int_{-\rho_V(M)}^{\rho_V(M)} \!
            \Phi(\xi)\xi^m \mathrm{d}\xi, \label{eq_phi_v_intermediate_expansion}
\end{equation}
where the exchange of the summation is permitted since
the integral is bounded to the convergent region $(-\rho_V(M), \rho_V(M))$.
Solving the remaining integral yields an approximation in the desired form of
(\ref{eq_alpha_expansion}). By noting the parts of (\ref{eq_phi_v_integral})
left out by the adjustments an exact expression for the
remainder $H_M^{\{\varphi\}}(v)$ is revealed
\begin{equation}
\begin{split}
    &H_M^{\{\varphi\}}(v) = \
        \int_{-\rho_V(M)}^{\rho_V(M)} \!
        \Phi(\xi)\Upsilon_M^{\{f\}} \mathrm{d}\xi \\
        &\quad + 
        \int_{-\infty}^{-\rho_V(M)} \!
        \Phi(\xi)f(\xi v) \mathrm{d}\xi
        \, + 
        \int_{\rho_V(M)}^{\infty} \!
        \Phi(\xi)f(\xi v) \mathrm{d}\xi \label{eq_general_remainder}
\end{split}
\end{equation}
such that the expansion and its coefficients are given as
\begin{align}
\begin{split}
    \label{eq_alpha_expansion_full}
    \varphi(v) &\simeq \sum_{m=0}^{M} \alpha_m(M) v^m \ \ \forall \ \ |v| < V,
\end{split}
\\
\begin{split}
    \alpha_m(M) &= \tilde{f}(m)
    \int_{-\rho_V(M)}^{\rho_V(M)} \!
    \Phi(\xi)\xi^m \mathrm{d}\xi . \label{eq_alpha_coef_general_form}
\end{split}
\end{align}

\subsection{Alternate Taylor Series Derivation}
We derive the Taylor series of $\tanh(v)$ by expressing
it in terms of its Fourier transform. Identities from
\cite{bateman_integral} yield
\begin{align}
    \tanh(v) &= \lim_{\rho \to \infty}
    \int_0^\rho \!
        \csch\left(\frac{\pi}{2}\xi\right)\sin(\xi v) \,
    \mathrm{d}\xi.
\end{align}
Expanding the kernel $\sin(\xi v)$ to a partial sum yields
\begin{equation}
\begin{split}
    & \tanh(v) \simeq \\
        &\ \lim_{\rho \to \infty}
        \sum_{m=1}^{M}
        \frac{(-1)^{m-1} v^{2 m - 1}}{(2 m - 1)!}
        \int_0^\rho \!
            \xi^{2 m - 1}\csch\left(\frac{\pi}{2}\xi\right)
        \mathrm{d}\xi.  \label{eq_tanh_intermediate_taylor}
\end{split}
\end{equation}
Integrating yields the expansion and exact error, $H_M(v)$:
\begin{align}
    \tanh(v) \simeq
    \sum_{m=1}^{M}\frac{2^{2m}(2^{2m}-1)B_{2m}v^{2m-1}}{(2m)!},
                      \ \forall \ |v| < \frac{\pi}{2},
\end{align}
\begin{align}
    H_{M}(v) =
        \frac{(-1)^Mv^{L}}{L!}
        \int_0^\infty \!
            \xi^{L}\csch\left(\frac{\pi}{2}\xi\right)
            \! {}_{1}F_{2}
            \left(1; b_1, b_2; z\right)
        \mathrm{d}\xi.
\end{align}

Here $L = 2M+1$, $b_1 = M+1$, $b_2 = M+3/2$, and $z = -\frac{v^2\xi^2}{4}$.
Noting that the manipulations of (\ref{eq_tanh_intermediate_taylor})
are only valid within the convergent regions of the Taylor
expanded kernel $\sin(\xi v)$ illuminates the source of
divergence for $|v|\geq \pi/2$ which is then mitigated via the AMITE
modifications.

\subsection{Modifying the Limits of Integration}

To modify the integral of Eq. (\ref{eq_tanh_intermediate_taylor})
so that the resulting expansions are valid for $|v| \geq \frac{\pi}{2}$ we require
an upper bound $\rho$ on the domain of validity
of the $M$-term partial sum expanded kernel. Following section
\ref{section_method_technique} we factor the partial sum remainders
to consider the dominant factors contributing to
divergence. We treat $\sin(\xi v)$ and $\cos(\xi v)$ which
are employed subsequently to expand $\tanh(v)$ and $\relu(v)$
as examples illustrating use of the technique on odd and
even nonlinearities. We begin with the partial sums
\begin{align}
    \cos(u) &= 
        \sum_{m=0}^{M}\frac{(-1)^m u^{2 m}}{(2 m)!}
      + \Upsilon_M^{\{\cos\}}(u), \label{eq_partial_sum_cos} \\
    \sin(u) &= 
        \sum_{m=0}^{M}\frac{(-1)^m u^{2 m + 1}}{(2 m + 1)!}
      + \Upsilon_M^{\{\sin\}}(u) \label{eq_partial_sum_sin}.
\end{align}
Factoring the remainders in the form of
(\ref{eq_remainder_factorization}) yields
\begin{align}
\begin{split}
    & \Upsilon_M^{\{\cos\}}(u) = \\
    & \quad
        \frac{(-1)^{M+1} u^{2M+2}}{(2M + 2)!}
        \ {}_{1}F_{2}
        \left(1; M+2, M+\frac{3}{2}; -\frac{u^2}{4}\right), 
        \label{eq_cos_remainder_hypergemom}
\end{split}
\\
\begin{split}
    & \Upsilon_M^{\{\sin\}}(u) = \\
    & \quad
        \frac{(-1)^{M+1} u^{2M+3}}{(2M+3)!}
        \ {}_{1}F_{2}
        \left(1; M+2, M+\frac{5}{2}; -\frac{u^2}{4}\right)
        \label{eq_sin_remainder_hypergeom}.
\end{split}
\end{align}

As $u$ tends to infinity the hypergeometric factors in
(\ref{eq_cos_remainder_hypergemom}) and (\ref{eq_sin_remainder_hypergeom})
tend to zero and the outer factors
\begin{equation}
Q_M^{\{\cos\}}(u) = \frac{u^{2M+2}}{(2M+2)!}, 
\quad Q_M^{\{\sin\}}(u) = \frac{u^{2M+3}}{(2M+3)!},
\end{equation}
tend to infinity. Noting that the outer factors drive the remainders'
growth we define the range of validity of the partial sums
in (\ref{eq_partial_sum_cos}) and (\ref{eq_partial_sum_sin})
to be all $u$ such that
$Q_M^{\{\cos\}}(u) < 1$ and $Q_M^{\{\sin\}}(u) < 1$
respectively.  This yields the required bounds on the
domain of validity of each partial sum:
\begin{align}
    \cos(u) &\simeq \sum_{m=0}^M \frac{(-1)^m u^{2m}}{(2m)!}
        \ \forall\ u < ((2M+2)!)^{\frac{1}{2M+2}}, \\
    \sin(u) &\simeq \sum_{m=0}^M \frac{(-1)^m u^{2m+1}}{(2m+1)!}  
        \ \forall\ u < ((2M+3)!)^{\frac{1}{2M+3}}.
\end{align}

Noting that the argument to the kernel in (\ref{eq_tanh_intermediate_taylor})
is $u = \xi v$, and adjusting for the maximum value $V = \max(v)$, we
let $\sigma$ and $\tau$ denote the required upper bounds for
integrals in the forms $\int \Phi(\xi)\cos(\xi v)\mathrm{d}\xi$ and
$\int \Phi(\xi)\sin(\xi v)\mathrm{d}\xi$
respectively:
\begin{align}
    \sigma = \frac{((2M+2)!)^{\frac{1}{2M+2}}}{V}, \ \ 
    \tau = \frac{((2M+3)!)^{\frac{1}{2M+3}}}{V}.
    \label{eq_range_of_validity}
\end{align}
\subsection{Hyperbolic Tangent Expansion}
Having identified the bound, $\tau$, for integrals having
kernels $\sin(\xi v)$ we can treat the hyperbolic tangent as
an exemplary odd nonlinearity.  Returning to
(\ref{eq_tanh_intermediate_taylor}) and integrating the component
$
\Lambda_j(\xi) = 
    \int \!
        \xi^j\csch\left(\frac{\pi}{2}\xi\right)
    \mathrm{d}\xi \label{eq_xi_csch_integral}
$
from $0$ to $\tau$ yields the formulas for the
coefficients $T_{2m + 1}(M)$ in (\ref{eq_tanh_approx_form}).
\begin{align}
\begin{split}
    \label{eq_tanh_expansion_full}
    \tanh(v) &\simeq \sum_{m=0}^{M} T_{2m+1}(M) v^{2m+1} \ \forall \ |v| < V,
\end{split}
\\[1ex]
\begin{split}
    T_{2m+1}(M) &= \frac{2^{2m+1}(4^{m+1}-1)B_{2m+2}}{(m+1)(2m+1)!} \\
    \quad + &\frac{(-1)^{m+1}4}{\pi} \sum_{k=0}^{2m+1}
        \frac{2^k \tau^{2m+1-k}
        \chi_{k+1}\left(e^{-\frac{\pi \tau}{2}}\right)}
        {\pi^{k}(2m+1-k)!}.
        \label{eq_xi_csch_solution}
\end{split}
\end{align}
Here, $\chi_s(z) = \frac{1}{2}(\polylog_s(z) - \polylog_s(-z))$ denotes
Legendre's Chi function.  (The result of (\ref{eq_xi_csch_solution}) is
derived fully in Appendix \ref{inegration_identity}.) The precise
remainder follows directly from application of
(\ref{eq_general_remainder}):
\begin{align}
\begin{split}
    &H_M^{\{\tanh\}}(v) = \int_\tau^{\infty} \!
        \csch\left(\frac{\pi}{2}\xi\right)
        \sin(\xi v) \mathrm{d}\xi
\\
    & \ + \frac{(-1)^{M+1}v^L}{L!} \int_0^{\tau} \!
        \xi^L \csch\left(\frac{\pi}{2}\xi\right)
        {}_{1}F_{2}(1; b_1, b_2; z) \mathrm{d}\xi, \label{eq_tanh_remainder1}
\end{split}
\\
\begin{split}
    L = 2M+3, b_1 = M+2, b_2 = M+\frac{5}{2}, z = -\frac{v^2 \xi^2}{4}.
\end{split}
\end{align}

The second term (\ref{eq_tanh_remainder1}) dominates for
large $v$.  Since it follows
from the Taylor remainder $\Upsilon_M^{\{\sin\}}(\xi v)$ it is
small for $v$ within the region of convergence.
Meanwhile, the first term in (\ref{eq_tanh_remainder1})
dominates for small $v$ leading to Gibbs-like oscillation of the
error for $|v| < V$.  Noting this, the first term in
(\ref{eq_tanh_remainder1}) can be integrated to analyze
errors inside the region of convergence.
Let $I_M^{\{\tanh\}}(v)$ denote this error component:
\begin{gather}
\begin{split}
    \label{eq_beta_fn_error_part}
    &I_M^{\{\tanh\}}(v) = \int_\tau^{\infty} \!
        \csch\left(\frac{\pi}{2}\xi\right)
        \sin(\xi v) \mathrm{d}\xi \\
        &\quad  = 2 \tanh(v) - \frac{i}{\pi}\left(
            \mathrm{B}_{e^{\pi \tau}}\left(a, 0 \right)
            - \mathrm{B}_{e^{\pi \tau}}\left(a^*, 0 \right)\right), \\
\end{split}
\end{gather}
Here, $\mathrm{B}_z(a,b)$ denotes the Incomplete Beta function
and $a = \frac{1}{2} + \frac{iv}{\pi}$.
Rearranging (\ref{eq_beta_fn_error_part}) to extract the
dominant oscillating factor, noting that for $a = \frac{1}{2} \pm \frac{iv}{\pi}$
\begin{align}
\begin{split}
    \mathrm{B}_{e^{\pi \tau}}\left(a, 0 \right) =
    \frac{e^{\frac{\pi\tau}{2}} e^{\pm i \tau v}}{a}
    {}_{2}F_{1}\left(1, a;
        \frac{3}{2} \pm \frac{i v}{\pi};
        e^{\pi \tau}\right),
\label{eq_beta_hypergeom_factor}
\end{split}
\end{align}
and substituting this result (\ref{eq_beta_hypergeom_factor}) into
(\ref{eq_beta_fn_error_part}) reveals
\begin{gather}
    I_M^{\{\tanh\}}(v) = 
        \frac{e^{\frac{\tau \pi}{2}}}{e^{\pi \tau} - 1}
        \left(U F + U^* F^*\right),
\label{eq_tanh_error_im_exact}
\\
U = \frac{e^{-i \tau \pi}}{v - \frac{i \pi}{2}}, \ 
F = {}_{2}F_{1}\left(1, 1;
        \frac{3}{2} + \frac{i v}{\pi};
        \frac{1}{1 - e^{\pi \tau}}\right).
\end{gather}
Recalling from (\ref{eq_range_of_validity}) that $\tau$ increases
with the number of terms $M$ and noting
$\lim_{\tau \to \infty} {}_{2}F_{1}\left(1, 1; c; \frac{1}{1-e^{\pi \tau}}\right) = 1$
reveals a remarkably accurate approximate closed form for the error within
the convergent region $|v| < V$:
\begin{align}
I_M^{\{\tanh\}}(v) \simeq
\frac{
4 e^{\frac{\pi \tau}{2}}
\left(\pi \sin(\tau v) + 2 v \cos(\tau v)\right)
}{(e^{\pi \tau} - 1)(4v^2 + \pi^2)}
\label{eq_tanh_approx_error_convergent_region}
\end{align}
From (\ref{eq_tanh_approx_error_convergent_region}) we can deduce
that within the convergent region the error will oscillate
with approximate period of $\frac{2 \pi}{\tau}$ and amplitude
dependent on $v$ and $\tau$.  Since $\tau$ increases with respect to $M$ and
$\lim_{\tau \to \infty} e^{\frac{\pi \tau}{2}}(e^{\pi \tau} - 1)^{-1} = 0$
we deduce from the exact expression (\ref{eq_tanh_error_im_exact}) for $I_M^{\{\tanh\}}(v)$ that
$\lim_{M \to \infty} I_M^{\{\tanh\}}(v) = 0$ as is
required to achieve a low error within $(-V, V)$.

\subsection{An Improved ReLU Expansion}
Here we expand $\relu(v)$, again following
Section \ref{section_method_technique}.
Since the odd part, $v$, of $\relu(v)$ is trivial, we
begin by expressing the
even part in terms of its Fourier transform
\begin{equation}
    |v| = - \frac{1}{\pi} \lim_{\rho \to \infty} 
    \int_{-\rho}^{\rho} \! \frac{\cos(\xi v)}{\xi^2} \mathrm{d}\xi.
\end{equation}
We then apply our bound $\sigma$ for integrals involving $\cos(v \xi)$
as per (\ref{eq_range_of_validity}) so that we may expand the kernel as a
partial sum and exchange the order of summation as per 
(\ref{eq_phi_v_intermediate_expansion}) yielding
\begin{equation}
    |v| \simeq -\frac{1}{\pi} \sum_{m=0}^{M}
    \frac{(-1)^m v^{2m}}{(2m)!}
    \int_{-\sigma}^{\sigma} \! \xi^{2m-2} \mathrm{d}\xi.
\end{equation}
Solving directly and adding back in the odd part of $\relu(v)$ yields the desired
formulas in Eq. (\ref{eq_relu_approx_form}).
\begin{gather}
\begin{split}
    \label{eq_relu_expansion_full}
    \relu(v) &\simeq \frac{v}{2} +\sum_{m=0}^{M} R_{2m}(M) v^{2m} \ \forall \ |v| < V,
\end{split}
\\[1ex]
\begin{split}
    R_{2m}(M) = \frac{(-1)^{m+1} \sigma^{2m-1}}{\pi (2m)! (2m - 1)}.
        \label{eq_relu_solution}
\end{split}
\end{gather}
Again, the precise remainder follows from (\ref{eq_general_remainder}):
\begin{align}
\begin{split}
    &H_M^{\{\relu\}}(v) = -\frac{1}{\pi} \int_\sigma^{\infty} \!
        \frac{\cos(\xi v)}{\xi^2} \mathrm{d}\xi
\\
    &\quad + \frac{(-1)^M v^{2M+2}}{\pi (2M+2)!} \int_0^{\sigma} \!
        \xi^{2M}{}_{1}F_{2}(1; b_1, b_2; z) \mathrm{d}\xi,
\end{split}
\\
\begin{split}
    &\quad\ b_1 = M+2, b_2 = M+\frac{3}{2}, z = -\frac{v^2 \xi^2}{4}.
\end{split}
\end{align}
The integrals can be readily solved yielding
\begin{align}
\begin{split}
    &H_M^{\{\relu\}}(v) = 
    \frac{|v|}{2} - \frac{v \mathrm{Si}(\sigma v)}{\pi}
        - \frac{\cos(\sigma v)}{\sigma}
\\
    &\quad\quad+\ \frac{(-1)^M \sigma^{2M+1} v^{2M+1}}{\pi (2M+2)(2M+1)!}
    {}_{2}F_{3}\left(1, a; b, b, c; z\right) \label{eq_relu_remainder1},
\end{split}
\\
\begin{split}
    a = M + \frac{1}{2}, b = M + \frac{3}{2}, c = M + 2, z = -\frac{\sigma^2 v^2}{4}.  \label{eq_relu_remainder3}
\end{split}
\end{align}

Due to the $v^{2M+1}$ factor, the second term of
(\ref{eq_relu_remainder1}) dominates outside the region
of convergence. Meanwhile, the first term dominates
for small $v$ leading to Gibbs-like oscillation within the
region of convergence $|v| < V$.  The exact expression for
the oscillating component of the error, $I_M^{\{\relu\}}(v)$,
is given by
\begin{align}
I_M^{\{\relu\}}(v) = \frac{|v|}{2} - \frac{v \mathrm{Si}(\sigma v)}{\pi}
        - \frac{\cos(\sigma v)}{\sigma}.
\end{align}
As with $\tanh(v)$, since $\sigma$ increases with respect to the number of terms, $M$,
and then since $\lim_{\sigma \to \infty}I_M^{\{\relu\}}(v) = 0$ we can again
deduce that $\lim_{M \to \infty}I_M^{\{\relu\}}(v) = 0$ as is required
to achieve low error within the convergent region.

\subsection{Full Layer Expansion}
\label{section_full_single_layer_expansion}
The coefficients, $\alpha_j$, of the activations, $\varphi$,
can then be used to expand a full network layer as a polynomial.
To perform this expansion, we substitute
the hidden neuron input as defined in Eq. (\ref{eq_hidden_neuron_input})
into Eq. (\ref{eq_alpha_expansion}).
\begin{equation}
    \varphi(v_n) \simeq
    \sum_{m}
    \alpha_j \left(b_n^{\{0\}} + \sum_{i=1}^{N_I} w_{n,i}^{\{0\}}x_i \right)^j,
\end{equation}
Here $j$ is the exponent belonging to each index $m$. Applying the multinomial expansion gives
\begin{align}
    \varphi(v_n) \simeq \sum_{m}
    \sum_{|\kappa|=j} \alpha_j \begin{pmatrix} j \\[-6pt] \kappa \end{pmatrix}
    \vartheta_n^{\kappa} x^{\kappa}, \\
    \vartheta_n = \begin{pmatrix}
    b_n^{\{0\}} & w_{n,1}^{\{0\}} & \cdots & w_{n,N_I}^{\{0\}}
    \end{pmatrix}.
\end{align}
Here
$\kappa =
\left[
    \kappa_1, \kappa_2, \cdots, \kappa_{N_I+1}
\right]$
is a multi-index where
$u^\kappa = \prod_{i} u_i^{\kappa_i}$, $|\kappa| = \sum_i \kappa_i$,
and
$\binom{j}{\kappa} = 
\frac{j!}{\kappa_1!\kappa_2! \cdots \kappa_N!}
$
is a multinomial coefficient.  The output of the layer can then
be approximated by taking the weighted sum of the neuron outputs:
\begin{equation}
\label{eq_network_output_full_expansion}
    y(\alpha, \beta, x) \simeq 
    b^{\{1\}} + 
    \sum_{n=1}^{N_H}
    \sum_{m}
    \sum_{|\kappa|=j}
    \alpha_j(M) \binom{j}{\kappa}
    w_n^{\{1\}} \vartheta^{\kappa} x^{\kappa}
\end{equation}
We use $\Psi_{\kappa}(\alpha, \beta)$ to denote the coefficient
associated with the term
$x^{\kappa} =
x_1^{\kappa_1}x_2^{\kappa_2} \cdots x_{N_I+1}^{\kappa_{N_I+1}}
$
in the polynomial approximation of $y(\alpha, \beta, x)$.  By definition
of $\Psi_{\kappa}(\alpha, \beta)$ we can write:
\begin{equation}
\label{eq_psi_coef_representation}
    y(\alpha, \beta, x) \simeq
        \sum_{\kappa} \Psi_{\kappa}(\alpha, \beta) x^{\kappa}
\end{equation}

\section{Expansion Results}
\label{section_expansion_results}
To verify the analysis of Section \ref{section_improved_expansions}
we evaluate the expansions for $\tanh$ and $\relu$.
In each evaluation, we first compare the approximation values
$\varphi_a(v) \simeq \sum_{m=0}^M\alpha_m v^m$ to the true
function values $\varphi(v)$ over $v$ in $(-V - c, V + c)$
where $c$ is small value chosen to show performance just outside
of the domain of validity $(-V, V)$.  We then show the
measured error
$E_M^{\{\varphi\}}(v) = \varphi(v) - \varphi_a(v)$,
exact error $H_M^{\{\varphi\}}(v)$, and analytic
approximate error $I_M^{\varphi}(v)$.  Finally, we display
the errors between the error formulas themselves and
the measured errors,
$\left|E_M^{\{\varphi\}}(v) - H_M^{\{\varphi\}}(v)\right|$ and
$\left|E_M^{\{\varphi\}}(v) - I_M^{\{\varphi\}}(v)\right|$.

\begin{figure}[p!]
    \centering
    \textit{
    Hyperbolic Tangent Expansion for $M=25$, $V=20$
    }\par\medskip
    \includegraphics[width=3.4in]{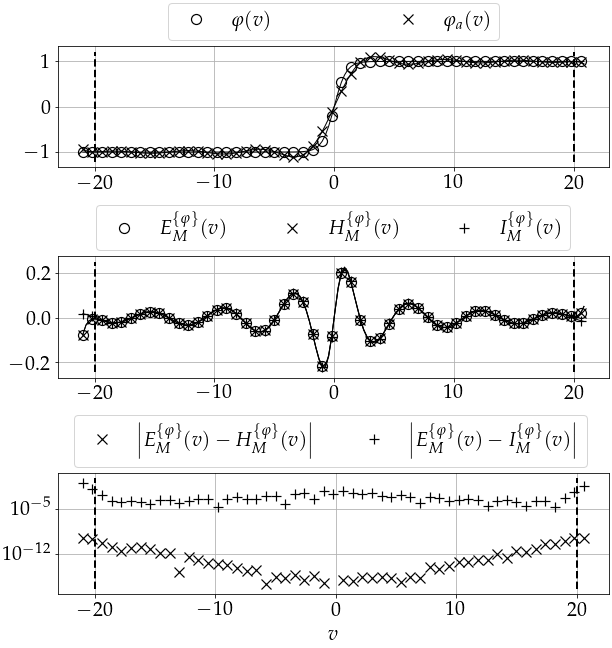}
    \caption{Expansion results and errors for $\varphi(v) = \tanh(v)$,
    number of terms  $M=25$ and validity parameter $V=20$.
    Top: expansion $\varphi_a(v)$ and true function $\varphi(v)$;
    middle: measured error $E_M^{\{\varphi\}}(v)$, exact error
    $H_M^{\{\varphi\}}(v)$, approximate error $I_M^{\{\varphi\}}(v)$;
    bottom: difference between measured errors and predicted errors.
    }
    \label{fig_tanh_big_m_25_v_max_20}
\end{figure}

\begin{figure}[p!]
    \centering
    \textit{
    ReLU Expansion for $M=25$, $V=20$
    }\par\medskip
    \includegraphics[width=3.4in]{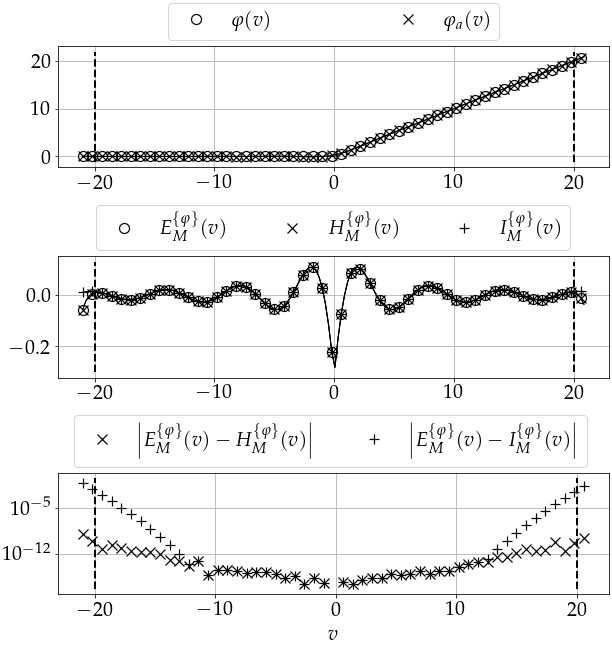}
    \caption{Expansion results and errors for $\varphi(v) = \relu(v)$,
    number of terms  $M=25$ and validity parameter $V=20$.
    Top: expansion $\varphi_a(v)$ and true function $\varphi(v)$;
    middle: measured error $E_M^{\{\varphi\}}(v)$, exact error
    $H_M^{\{\varphi\}}(v)$, approximate error $I_M^{\{\varphi\}}(v)$;
    bottom: difference between measured errors and predicted errors.
    }
    \label{fig_relu_big_m_25_v_max_20}
\end{figure}

Expansions and errors are shown in 
Fig. \ref{fig_tanh_big_m_25_v_max_20} and
Fig. \ref{fig_relu_big_m_25_v_max_20}.  Formulas to
compute coefficients are evaluated in 450 digits of
precision and the polynomials themselves are evaluated
in IEEE 754 double precision floating point arithmetic.
Satisfaction of the qualitative properties described
in Table \ref{table_expansion_properties_considered}
is summarized in Table \ref{table_qualitative_comparison}.
We find that the \textit{exact coeffs.},
\textit{monic form}, \textit{adjustable domain}, and
\textit{handles undef. derivs.} properties are satisfied
by (\ref{eq_tanh_expansion_full}) and
(\ref{eq_relu_expansion_full}).  The \textit{explicit exact errors}
property is satisfied by (\ref{eq_tanh_remainder1}) and
(\ref{eq_tanh_remainder1}) and
(\ref{eq_relu_remainder1}).
The \textit{consistent method} property is
satisfied by Section \ref{section_method_technique}.

\section{Case Study: Applying AMITE to MLP Equivalence Testing}
\label{section_network_replication}
\subsection{Definitions}
In this case study, we employ the following definitions.
An \emph{original network} is a neural network,
created by a designer, e.g., in software.
A \emph{network under test} is an implemented
neural network (e.g., in hardware) which a
test procedure must determine to be output-equivalent to the original
network.  The \emph{stimulus signal} is the signal applied and the
\emph{response signal} is the output signal measured, including
measurement noise.  A \emph{replicated network} is defined as the
network having weights extracted from the network under test.
A network under test is confirmed to be \emph{output-equivalent} to the
original network if it is shown that they will produce sufficiently
close outputs for all inputs in a given interval, according to a
provided error metric.
\subsection{Experimental Setup}
\label{section_experimental_setup}
The experimental process was modeled computationally on a
Dell XPS Laptop running Python 3.7,
using the pytorch, numpy, cupy, and mpmath libraries, on an
Nvidia GeForce GTX 1650 GPU with 1024 Cuda Cores and an
Intel i9-9980HK CPU at 2.40GHz.  Parallel processing
was implemented through pytorch and cupy matrix operations.
Coefficients for $\tanh$ and $\relu$ were precomputed using 450 digits
of precision via the precise formulas of Section \ref{section_improved_expansions}
and polynomials were evaluated at runtime in IEEE 754 double precision floating
point arithmetic.  Mathematical primitives including binomial and multinomial coefficients
were precomputed and stored in hashmaps.  Multinomial coefficient
hashmaps required 4.73GB of disk space and were loaded into
program memory as needed in chunks ranging from 11KB to 986MB
for fast access.

\subsection{Example Application to Equivalence Testing}
As a first example, we consider a single hidden layer MLP
used to control a fielded hardware system. In such
applications, it can be advantageous to rapidly develop
models via open-source software
\cite{scikit-learn,pytorch,tensorflow}
and then exploit the efficiency of numerous proposed
hardware approaches for implementation
\cite{misra_artificial_2010,maliuk_experimentation_2015,abdelsalam_accurate_2016,schuman_survey_2017}.
However, separating design from implementation
motivates the question: how does one confirm that the
network implemented in hardware is output-equivalent
to the network designed?
The answer is complicated by equivalent
configurations of a given network
\cite{sussmann_uniqueness_1992,albertini_uniqueness_1993}
and by small parameter perturbations
yielding approximately equivalent input-output pairs
\cite{dimattina_how_2010}. Measurement noise obscures the true error
between the implemented input-output map and the original
input-output map and due to nonlinearity,
it is unclear what degree of deviation represents a
defect.

To show how AMITE can be applied here, we develop a
bug-finding ``conformance test'' to address the challenges above.
First, we extract the weights of the implemented network
following the methods of model stealing / replication
\cite{tramer_stealing_2016,papernot_practical_2017,oh_reverse_2019,duddu_stealing_2018,orekondy_knockoff_2019} to fit a new model
to a set of stimulus-response pairs.  Then following the
methods of Section \ref{section_improved_expansions}
we analytically convert the weights of a single network layer
to an AMITE polynomial and compare the coefficients to the AMITE
representation of the original network.
Here the AMITE coefficients provide a means to assess equivalence directly
from the replicated and original weights.

The preceding replication
step is used to provide a guarantee that the stimulus applied
yields suitable coverage to achieve a comprehensive equivalency
test.  If the test signal does not provide sufficient coverage
invalid weights will be extracted and the test will fail.
This step biases the test towards a low false negative
rate while offering the opportunity to further inspect a
failed result by adding more test inputs. The observed
efficacy of the equivalency test is dependent on the
expansion technique's ability to accurately and efficiently
expand an entire network layer as a polynomial over a desired domain.
This first case study assumes low dimensional, normalized
inputs and assumes the expected structure of the network under test is known
while its actual structure is unknown.

\subsection{MLP Equivalence Test Steps}
\label{section_equivalency_test_steps}
The specific steps of the equivalency test used are as
follows. First, an interrogation signal, $x_I(t)$,
consisting of white noise fuzz vectors
is activated to stimulate the implemented network.
Measured output with any incurred
noise $\mathcal{N}$ is collected and denoted
$\hat{y}_I(t) = y\left(\alpha, \beta_I, x_I(t)\right) + \mathcal{N}.$
The input-output pairs $\left(x_I, y_M\right)$ are
then used to solve for the replicated network weights
$\beta_{R}$ by fitting:
\begin{equation}
\label{eq_optimization_for_replicated_network}
\beta_R = \
    \underset{\beta}{\arg\min}\left(
        \left(\hat{y}_I(t) - y\left(\alpha, \beta, x_I(t)\right)\right)^2
    \right)
\end{equation}

Since the expected $\beta$ values are known in advance, they are
used as a starting point for fitting to expedite extraction of $\beta_R$
and avoid false positives due to multiple local minima.
The optimization problem is solved over 250 fuzz vectors $x_I$
via stochastic gradient descent with adaptive learning rate
starting at $l=1\times10^{-3}$, weight decay $d=0.01$,
and no momentum. 
The original and replicated networks are then represented analytically
via the methods of Section \ref{section_full_single_layer_expansion}
as multivariate AMITE polynomials having coefficients
$\Psi(\alpha, \beta_O) = \left[
\Psi_{\kappa_n}(\alpha, \beta_O)
\right]$ and
$\Psi(\alpha, \beta_R) = \left[
\Psi_{\kappa_n}(\alpha, \beta_R)
\right]$ respectively.  To check equivalency, we use
a mean absolute logarithmic error between the magnitudes of the
coefficients:
\begin{equation}
\begin{split}
\label{eq_error_metric}
    & \eta(\Psi(\alpha, \beta_R), \Psi(\alpha, \beta_O)) = \\ 
    & \quad \mean\left|
        \log\left(\varepsilon + |\Psi(\alpha, \beta_O)|\right) -
        \log\left(\varepsilon + |\Psi(\alpha, \beta_R)|\right)
    \right|.
\end{split}
\end{equation}
If the error $\eta$ is below a set threshold $\epsilon$ the networks
are determined to be equivalent.  These steps are summarized in Fig.
\ref{fig_test_procedure}.

\subsection{Network Equivalency Testing Results}
\begin{figure}[!t]
\centering
\textit{Example Bug Detection Test}\par\medskip
\includegraphics[width=3.4in]{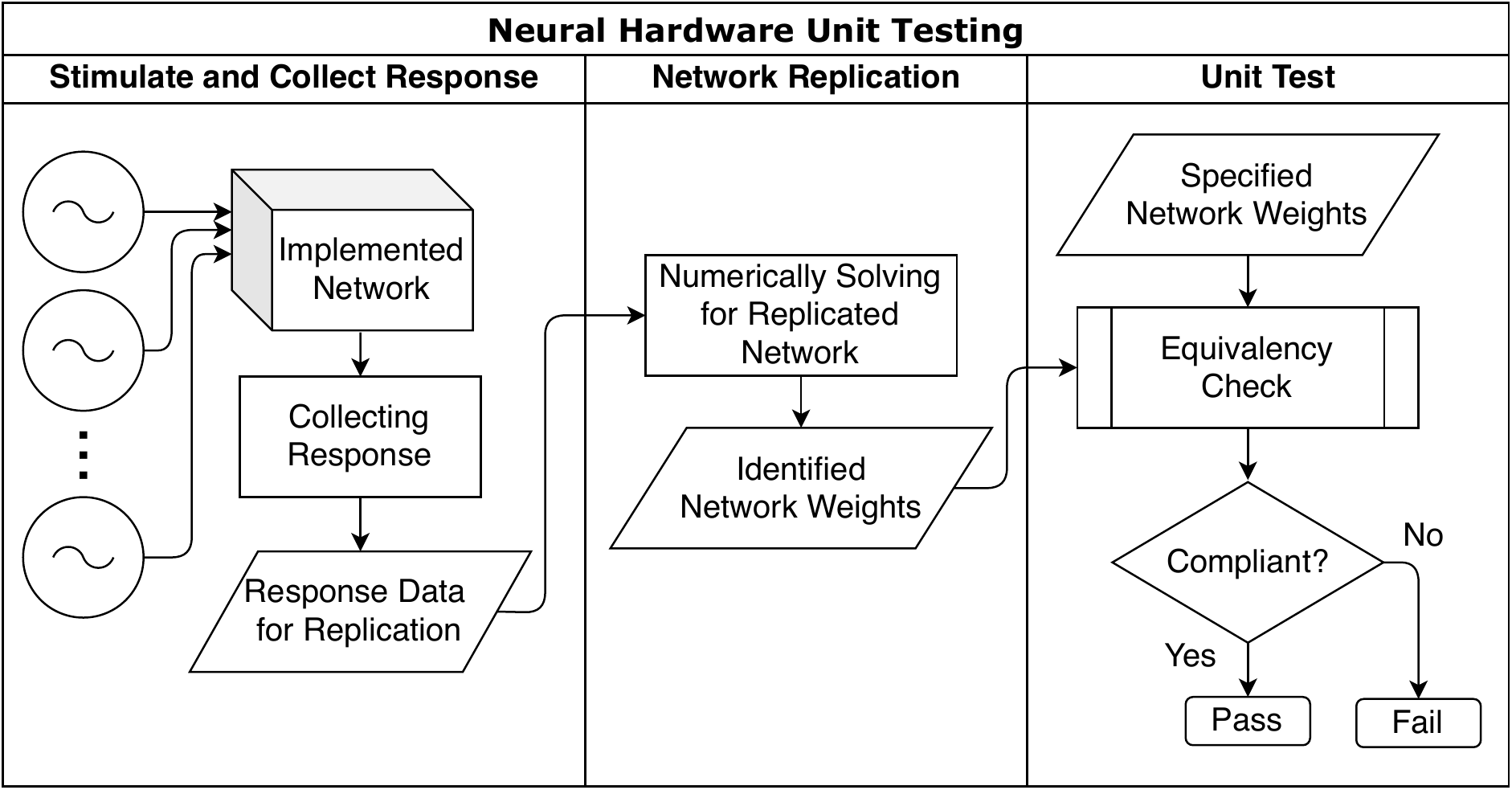}
\caption{Unit level bug-finding conformance test procedure applying
network replication to test an implemented MLP for equivalence to an
original MLP.  An implemented network of known expected structure with
unknown weights and unknown structure is stimulated by a plurality of
signals and the unknown weights are identified from the response.
The identified weights are converted to AMITE representations and
checked for equivalency to the specified weights to determine if the
networks match.
}
\label{fig_test_procedure}
\end{figure}

\begin{figure}[!t]
\centering
\textit{Bug Detection Accuracy wrt. Measurement Noise}\par\medskip
\includegraphics[width=3.4in]{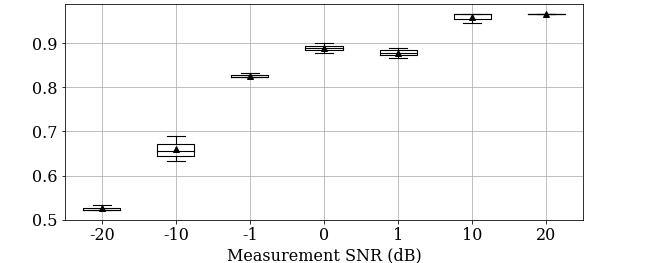}
\caption{Bug detection accuracy with respect to measurement noise.
The fraction of defective networks under test which were
correctly identified is shown here with respect to measurement noise.  Results are aggregated over 1890 calls
to the equivalency test on networks of varying sizes under
various levels of noise.
}
\label{fig_accuracy_wrt_snr}
\end{figure}

\begin{figure}[ht!]
\centering
\textit{Runtime (s) wrt. Number of Hidden Units, ($N_H$)}\par\medskip
\includegraphics[width=3.4in]{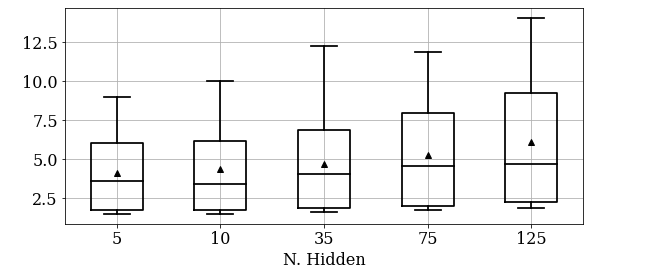}
\caption{Runtime per call to the test with respect to number of
hidden units, $N_H$.  Statistics are computed across 1890 calls to the
test.
}
\label{fig_runtime_wrt_hidden_units}
\end{figure}

\begin{figure}[ht!]
\centering
\textit{Runtime (s) wrt. Number of Inputs, ($N_I$)}\par\medskip
\includegraphics[width=3.4in]{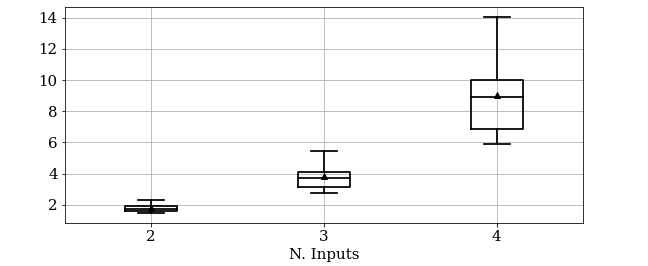}
\caption{Runtime per call to the test with respect to number of
inputs, $N_I$. Statistics are computed across 1890 calls to the
test.
}
\label{fig_runtime_wrt_num_inputs}
\end{figure}

To collect results for the method of section
\ref{section_equivalency_test_steps}, a population of 90 MLPs
is instantiated having inputs $N_I \in \{1, 2, 3\}$ and number of hidden nodes
$N_H \in \{5, 10, 35, 75, 125\}$.  The population is duplicated to
model a set of networks under test.  Defects in the form of random
weight perturbations are introduced into half of the population
of networks under test. The network equivalency test of
Fig. \ref{fig_test_procedure} is executed following the methods
of Section \ref{section_equivalency_test_steps} on each network.  All 90
networks are tested with varying noise
$\mathcal{N} \in \{-20, -10, -1, 0, 1, 10, 20\}$.  The entire procedure is
repeated three times for a total of 1890 calls to the test.
The accuracy at finding bugs is reported in 
Fig. \ref{fig_accuracy_wrt_snr} with respect to signal to noise
ratio in the measured outputs.  The runtimes with
respect to number of hidden nodes and number of inputs are reported in
Fig. \ref{fig_runtime_wrt_hidden_units} and
\ref{fig_runtime_wrt_num_inputs}.  For all
tests an error threshold of $\epsilon = 0.01$ was applied
when checking equivalency by the metric of (\ref{eq_error_metric}):
$
\eta(\Psi(\alpha, \beta_R), \Psi(\alpha, \beta_O)) > \epsilon
$.  We find that the test completes in under 7 seconds on average
for low-dimensional input (2-4 inputs) MLPs having up to 125
hidden nodes.  The runtime per test is inclusive of both solving
for the replicated network parameters and computing the AMITE
polynomial representation.


\section{Case Study: Range Bounding Deep FFNNs via AMITE Based Taylor Modeling}
\label{section_range_bounding}
\subsection{Experimental Setup}
The experimental setup is identical to the previous case study 
except for the digits of precision (DP) used to compute the
coefficients, which are reduced to 65 for the smaller networks and increased
up to 450 as needed.  A custom Taylor modeling library was implemented
in Python 3.7 allowing traditional Taylor series or AMITE polynomials
to be used in constructing Taylor models for comparison.
A subset of IEEE 1788.1 interval arithmetic
was implemented for handling error intervals.

\subsection{Example Application to Range Bounding Deep FFNNs}
This application is identical in setup and motivation
to the previous application, except that
we now desire formal bounds on the outputs of a neural network
which will control a fielded hardware system.  For example, if the network
is providing acceleration commands, it may be required to confirm
that the acceleration commands will never exceed certain values as long
as the input remains in a restricted domain.  The range analysis problem
has been studied in depth in the neural network literature and for
nonlinear systems in general.  A well-known tool for performing this
analysis is Taylor models \cite{berz_taylor_1997, berz_taylor_nodate},
however it is also known \cite{dutta_reachability_2019,dutta_output_2018}
that straightforward application of Taylor models to deep FFNNs
leads to over-approximation of errors.

In this example application, we address the problem of over-estimation of
errors analytically, employing the formulas of Section
\ref{section_improved_expansions} to inform the generation of Taylor
models for the networks under test, allowing efficient generation
of tighter range bounds for deep FFNNs over larger domain intervals than
possible with an unmodified Taylor model.  Use of such methods in critical
applications, such as \cite{pope_hierarchical_2021}, further motivates
the use of analytic methods in complement to the existing
repertoire of numerical methods for improving Taylor models.

\begin{figure*}[!tb]
\centering
\textit{Results for Range Bounding Deep FFNNs via Taylor Models Enhanced with AMITE Polynomials}\par\medskip
\includegraphics[width=\textwidth]{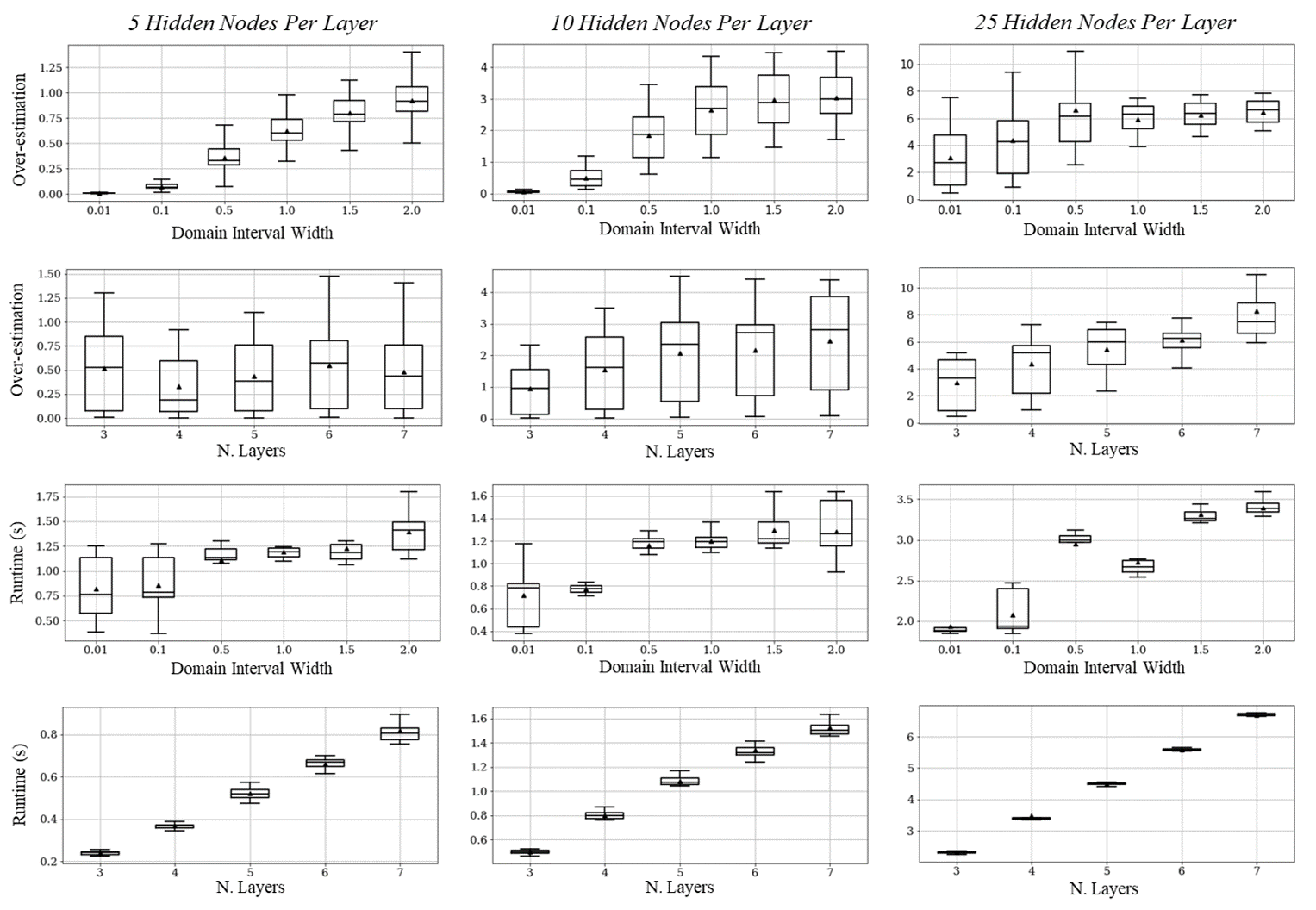}
\caption{Range bounding results over a population of deep FFNNs.  Here each
network in a population of deep FFNNs (3-7 layers and 5-25
hidden neurons per layer) is range bounded numerically via
random stimulation and rigorously via Taylor models employing AMITE polynomials.
In the top row, over-estimation of the Taylor models compared to the
numerical estimate is shown.  In the second row, over-estimation is shown with
respect to the number of hidden layers.  In the third row, the runtime
required to generate the AMITE polynomials (one time cost) is shown with
respect to the interval width the polynomials were optimized for. In the
bottom row, the runtime to range bound each network is shown with respect
to the number of hidden layers.
}
\label{fig_range_bounding_via_amite}
\end{figure*}

\begin{figure}[!tb]
\centering
\textit{Results for Range Bounding Deep FFNNs via
Traditional Taylor Models with Taylor Series}\par\medskip
\includegraphics[width=3.4in]{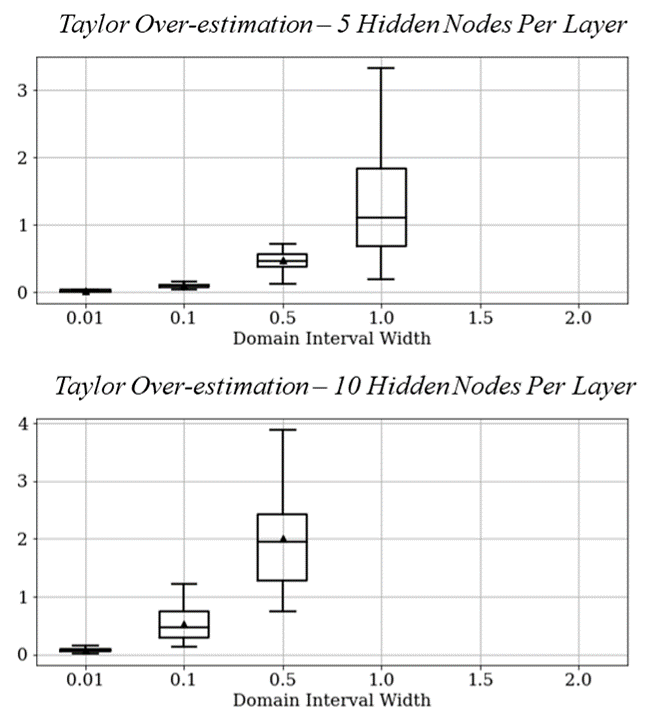}
\caption{Range bounding results over a population of deep FFNNs via
traditional Taylor models with Taylor Series.  Range bounds over a population
of deep FFNNs are show as computed via conventional Taylor models for domain
input intervals of increasing width.  The rapid expansion of the range interval
overestimation with respect to the input interval width is shown.  Note that
the conventional Taylor model diverges for inputs greater than $\frac{\pi}{2}$.
Propagation of errors through the network can cause the model to diverge sooner.
}
\label{fig_range_bounding_via_taylor}
\end{figure}

\subsection{Improved Taylor Modeling and Range Bounding Procedure}
The improved Taylor modeling and range bounding procedure for neural
networks enabled by the AMITE polynomials is described as follows.
First, the weights of the network and the desired domain intervals
$x_n \in \mathcal{I}_n$,
$\mathcal{I}_n = (x_{n,\mathrm{min}}, x_{n,\mathrm{max}})$ for
each input $x_n$ of the network are provided as procedure inputs.
The network to be range bounded is then stimulated via random white-noise
fuzz vectors in $\mathcal{I}_n$ to obtain a suitable initial guess for
the required domain of validity $V$.  A
conservative value $V = S\,\mathrm{max}(v)$ is chosen where
$S$ is a safety factor.  AMITE polynomials are computed for the
activation function given the detected $V$ and a desired number of terms $M$.
The AMITE polynomials are then used as the
coefficients $\left[\alpha_m\right]$ in the Taylor model.

To efficiently determine the error bounds on the Taylor model,
the minimum and maximum
error between the polynomial determined by $\left[\alpha_m\right]$ and the
true activation function $\varphi$ over $\mathcal{I}_n$ must be determined.
To do this efficiently, we first exploit the accurate simplified error
approximations, $I_M^{\{\varphi\}}(v)$,
derived in Section \ref{section_improved_expansions}.  Since
these formulas are efficient to compute, a course grid search over their range on
the given domain is performed to identify an initial estimate for the locations
of the maximum and minimum errors. These locations and the known periods of
oscillation in the errors, also derived in Section \ref{section_improved_expansions},
are employed to identify a bounded optimization region within one half period
in either direction of the initial estimate.

A bounded optimizer
is then employed to optimize over the exact error formula, $H_M^{\{\varphi\}}(v)$, and
find the precise maximum and minimum errors within a defined optimizer tolerance.
The optimizer tolerance is then added via interval addition to the error interval
to obtain a formal error bound on the expansion to be used in the Taylor model as
$(e_{\mathrm{min}}, e_{\mathrm{max}})$.  As such, a formal error bound
is given with a minimum number of expensive calls to evaluate the exact error
function. With a Taylor model,
$\mathcal{T} = \left(\left[\alpha_m\right], (e_{\mathrm{min}}, e_{\mathrm{max}})\right)$,
obtained for $\varphi(v)$, the rules of Taylor model and interval arithmetic
are then used to propagate the input domains $\mathcal{I}_n$ through the
network, adjusting the error interval after each operation.  If at any time
divergence is encountered (e.g., due to infinite results), the procedure
returns to the first step and increases the safety factor $S$ before the
Taylor model is reconstructed over a wider domain and the subsequent steps are
repeated.

\subsection{Range Bounding Results}
To collect results for range bounding example, a population of 60 deep FFNNs
is instantiated having 3 inputs, a uniform number of hidden nodes per layer
$N_H \in \{5, 10, 25\}$, and layers $N_L \in \{3,4,5,6,7\}$.
The range bounding procedure is then run on each network to analyze several
input interval widths.
For $N_H = 5, 10$, $M=6$ is used with 65 DP.  For
$N_H = 25$, $M=12$ is used with 130 DP.  For $N_H = 50$, $M=25$ is used
with 260 DP.  Each network is range bounded for input intervals centered
about zero having widths $W \in \{0.01, 0.1, 0.5, 1, 1.5, 2.0\}$.  Evaluating
15 networks over 6 intervals per network yields 90 calls to the test procedure per
network architecture for a total of 360 calls to the procedure. Each input
is set to the same domain interval for simplicity. For each network, its range
bounds are first estimated numerically
by stimulating the network with 100 random inputs inside the domain interval.
Then the range bounds are estimated via a traditional Taylor model and
via a Taylor model constructed from AMITE polynomials.

For each network, the
range overestimation via traditional Taylor models, range overestimation via
AMITE improved Taylor models, runtime to generate the AMITE coefficients,
and runtime to execute the range bounding procedure are tabulated.
The range overestimation is defined to be the width of the Taylor model interval
minus the width of the interval found by taking the maximum and minimum outputs
resulting from random stimulation.  Since the true range is unknown in all
cases, we note that the numerical estimate likely underestimates the true
range since random stimulation is not guaranteed to excite steep nonlinearities
in the network's input-output manifold.
Since $\relu$ does not have a conventional Taylor series,
a $\tanh$ activation function is used throughout
to enable direct comparison between conventional Taylor models
(via Taylor series) and Taylor models constructed from AMITE polynomials.
We note that conventional Taylor models diverged to infinity for many
cases tested.  While this could be resolved by subdividing the domain
into smaller intervals a single interval was employed in all cases
for a one-to-one comparison.

\begin{figure}[!tb]
\centering
\textit{Results for Range Bounding Wide MLPs via Taylor Models Enhanced with AMITE Polynomials}\par\medskip
\includegraphics[width=3.4in]{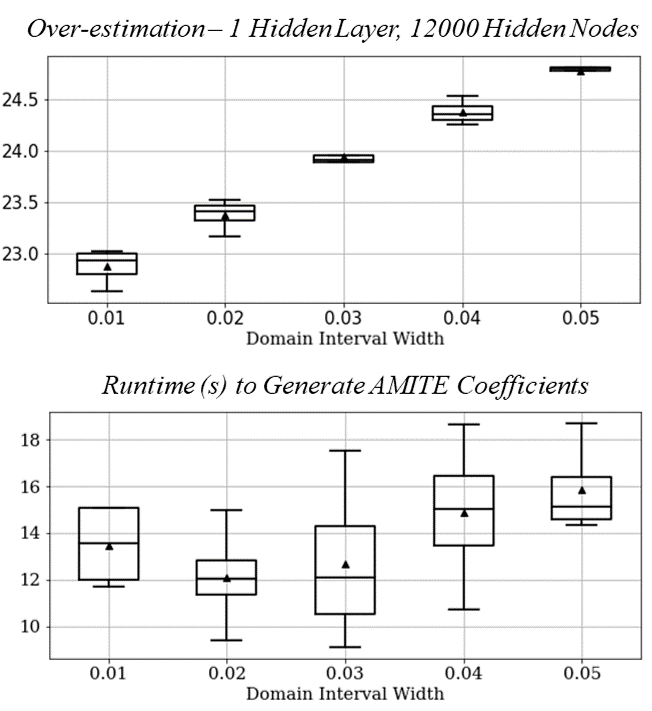}
\caption{Range bounding results for wide MLPs via Taylor models enhanced with AMITE polynomials.  Here range bounds for 4 MLPs, each having 1 hidden layer and 12000 hidden nodes are computed using AMITE enhanced Taylor models for several domiain interval widths and shown in the top plot.  Runtime (s) to generate the AMITE
polynomial coefficients is shown in the bottom plot with respect to the domain
interval width the coefficients were generated for.}
\label{fig_adf_networks}
\end{figure}

Results for using AMITE polynomials to construct Taylor models to range
bound deep FFNNs and overestimation compared to numerically estimated bounds
are shown in Fig. \ref{fig_range_bounding_via_amite}.  Results using
conventional Taylor modeling to attempt the same range bounding on
network architectures where this is feasible are shown in Fig. \ref{fig_range_bounding_via_taylor}.
In a final experiment to demonstrate a practical application,
a set of 4 wide networks of similar architecture
(1 hidden layer, 12K hidden nodes, $\relu$ activation) to those
which Pope et al. trained to defeat human pilots in simulated F-16
air-to-air combat \cite{pope_hierarchical_2021} is range bounded over increasingly wide input
intervals using AMITE polynomials.  Results
for this experiment are shown in Fig. \ref{fig_adf_networks}.  Runtime
to perform the range bounding was averaged 264 seconds per network.
Runtime to generate the AMITE coefficients optimized for each interval
size is between 9 and 19 seconds.  We note that due to the size of the
network and use of ReLU, a conventional Taylor model could not
be employed.

\section{Related Work}
\label{section_related_work}
This work is intended to apply
to several sub-fields of research
in neural networks, and is particularly related to
neural systems verification, equivalency testing,
and to model replication attacks. Recent work on
neural network verification focus on
defense against and identification of adversarial perturbations \cite{darvish_rouani_safe_2019,venzke_verification_2020}, output
reachable set estimation \cite{xiang_output_2018}, and computing
upper bounds on error rates under adversarial attack
\cite{dvijotham_efficient_2019}.  Liu et al. survey
algorithms for verifying deep neural networks and
categorize them into forms of reachability analysis, optimization
to falsify an assertion, and search to falsify an assertion
\cite{liu_Algorithms_2019}. Bunel et al. present a unifying
framework for the formal verification of piecewise linear
neural networks \cite{bunel_unified_2018}.

Other significant studies present output range analysis via
over-approximation \cite{gehr_ai2:_2018}\cite{huang_divide_2020}
and via exact (sound and complete) methods\cite{tran_nnv:_2020}
\cite{tran_parallelizable_2019}, verification against a
specification \cite{yaghoubi_worst-case_2020}, and a hybrid
systems approach to verifying properties of neural
network controllers with sigmoid activations\cite{ivanov_verisig:_2019}.
A survey of the safety and trustworthiness of deep neural
networks is provided by Huang et al. \cite{huang_survey_2020}.
A review of the state of the art in testing AI systems for safety
and robustness is provided by Wu et al. \cite{wu_testing_2020}.

Methods of identifying equivalent neural networks given
a matrix of neuron outputs are available
\cite{laakso_content_2000,li_convergent_2015,raghu_svcca_2017,morcos_insights_2018}
and have recently been advanced by Kornblith et al.
\cite{kornblith_similarity_2019}.
Methods to identify opportunities for compressing a neural
network \cite{chang_deep_2015,huang_ltnn_2019} and for
perturbing parameters to identify output-equivalent
neural networks are also available \cite{dimattina_how_2010}.

%It has recently been demonstrated that trained
%neural networks contain interpretable sub-circuits
%comprising semantic functions
%\cite{karpathy_visualizing_2015,zhou_object_2015,bau_network_2017,gonzalez-garcia_semantic_2017,olah2020zoom}.
%Building on feature visualization methods such as those from Erhan et al.
%\cite{erhan2009visualizing}, Simonyan et al.
%\cite{simonyan_deep_2014}, and Mordvintsev et al. \cite{mordvinsev2015inceptionism},
%a number of methods are now available to visualize
%individual sub-circuits inside neural networks 
%\cite{nguyen_synthesizing_2016,nguyen_plug_2017,olah2017feature,hohman_summit:_2019,carter2019activation}.  Furthermore, stacking networks such as
%\cite{deng_scalable_2012,deng_deep_2013,wang_svm-based_2019,khamparia_sound_2019}
%comprised of stacked smaller networks are good candidates
%for analysis via these techniques.
%We offer the methods of this paper as
%especially applicable to detailed study of
%such small sub-circuits within larger networks.

\section{Conclusion}
\label{section_conclusion}
In this paper we have developed a novel polynomial
expansion technique, the analytically modified
integral transform expansion (AMITE), to enable for the
first time decomposing neural network activation functions
into polynomial forms having six advantageous properties,
including exact formulas, monic form, adjustable domain,
and robustness to undefined derivatives. To complement the
exact formulas, convenient closed-form approximations of the
exact expansion errors have been developed by exploiting
relationships amongst the special functions arising therein.
We have demonstrated the effectiveness of AMITE in two case studies.
First, a method for decomposing a single hidden layer into
an multivariate polynomial is derived and applied to an
illustrative MLP equivalence testing problem where a black-box
network under test is stimulated, and a replicated multivariate
polynomial form is efficiently extracted from a noisy response
to enable comparison against an original network.  We then demonstrate
the effectiveness of AMITE for generating improved Taylor models for
deep FFNNs which allow efficiently obtaining tighter range bounds
over larger input intervals than possible via conventional Taylor models.
Our future work focuses on combining both case studies to efficiently
extract Taylor models of black box deep FFNNs from noisy stimulus-response
pairs, thus allowing range bounding of neural networks for which the user
has no access to the internal parameters.
%A key limitation is that runtime grows rapidly with the number of inputs.  As such, this technique is better suited for neural control applications with low dimensional inputs. Given that the approach can be efficiently applied, has several unique desirable properties, and given the use of polynomial approximations in neural network explainability, verification, and security, we conclude that the presented approach is a useful complement to the existing repertoire of polynomial expansions available for analysis of neural networks.

% use section* for acknowledgment
%\section*{Acknowledgment}

%The authors would like to thank...

% if have a single appendix:
%\appendix[Proof of the Zonklar Equations]
% or
%\appendix  % for no appendix heading
% do not use \section anymore after \appendix, only \section*
% is possibly needed

% use appendices with more than one appendix
% then use \section to start each appendix
% you must declare a \section before using any
% \subsection or using \label (\appendices by itself
% starts a section numbered zero.)
%

\appendices
\section{}
\label{inegration_identity}
Here we present a full derivation of the result of
eq. (\ref{eq_xi_csch_solution}) in section
\ref{section_improved_expansions}. This equation
gives the formula for the coefficients of the improved
$\tanh(v)$ expansion and follows from the integral
$
\int \!
    \xi^j\csch\left(\frac{\pi}{2}\xi\right)
\mathrm{d}\xi.
$
We solve this integral by starting with a trivially more
general one,
$
\int \! \xi^j \sech(a \xi + b) d\xi,
$
and exploiting the basic relation
$
\sech\left(a\xi + \frac{i \pi}{2}\right) = -i \csch(a\xi),
$
between the hyperbolic functions such that the result may be
readily applied to a class of related functions.  Starting
with repeated integration by parts
\begin{align}
\begin{split}
    &\int \! \xi^j \sech(a\xi + b) d\xi = \\
    &\ (-1)^j \int \! \frac{\partial^j}{\partial \xi^j} \xi^j \left(
    \underbrace{\int \! \mathrm{d\xi} \, \cdots \int \! \mathrm{d\xi}}_{j}
    \sech(a\xi + b) \right) \mathrm{d}\xi \\
    &\ + \sum_{k=0}^{j-1} (-1)^k \frac{\partial^k}{\partial \xi^k}\xi^{j}
    \frac{\partial^{j-k-1}}{\partial\xi^{j-k-1}}
    \underbrace{\int \! \mathrm{d\xi} \, \cdots \int \! \mathrm{d\xi}}_{j}
    \sech(a\xi + b).
    \label{eq_ibp_sech}
\end{split}
\end{align}
Repeated differentiation of the hyperbolics is presented
in full generality by \cite{qureshi_successive_2019}.  Here we
employ a simpler formula for repeated integration revealed by
expressing $\sech(a\xi + b)$ in terms of the polylogarithms:
\begin{align}
\begin{split}
&\underbrace{\int \! \mathrm{d\xi} \, \cdots \int \! \mathrm{d\xi}}_{j}
\sech(a\xi + b) = \\
&\quad \frac{i (-1)^j}{a^j}\left(
    \polylog_j\left(-i e^{-a\xi + b}\right)
    - \polylog_j\left(i e^{-a\xi - b}\right)
\right).
\end{split}
\end{align}
Applying to (\ref{eq_ibp_sech}) yields
the useful intermediate result
\begin{align}
\int \! \xi^j \sech(a\xi + b) \mathrm{d}\xi =
-\frac{2}{a} \sum_{k=0}^j \frac{j! \xi^{j-k}
\mathrm{Ti}_{k+1}(e^{-a\xi -b})}{a^k (j-k)!}.
\end{align}
Here $\mathrm{Ti}_s(z)$ denotes the Inverse Tangent Integral, defined by
\mbox{$
\mathrm{Ti}_s(z) = 
\frac{1}{2i}
\left( \polylog_s(i z) - \polylog_s(-i z) \right).
$}  Noting the relationship
\mbox{$i\mathrm{Ti}_s(-iz) = \chi_s(z)$} and the relationship
between $\csch(z)$ and $\sech(z)$ reveals
\begin{align}
\begin{split}
    \Lambda_j(\xi) &= \int \xi^j \csch(a \xi) \mathrm{d} \xi
\\
&= -\frac{2}{a}\sum_{k=0}^{j}
    \frac{j! \xi^{j-k} \chi_{k+1}(e^{-a\xi})}{a^k (j-k)!}.
\end{split}
\end{align}

The expression for the coefficient $T_{2m+1}(M)$ in
(\ref{eq_tanh_approx_form}) then follows directly from
(\ref{eq_alpha_coef_general_form}).  The integral is
evaluated to the bound $\tau$ corresponding to the kernel
$\sin(\xi v)$ such that
\begin{align}
T_{2m+1} &= \frac{(-1)^m}{(2m+1)!}
\int_0^{\tau} \xi^{2m+1} \csch\left(\frac{\pi}{2} \xi\right) \mathrm{d}\xi \\
&=\frac{(-1)^m}{(2m+1)!}\left(
\Lambda_{2m+1}(\tau) - \lim_{\varepsilon \to 0} \Lambda_{2m+1}(\varepsilon)
\right).
\end{align}
Noting that
\begin{align}
\begin{split}
    \frac{1}{2} \lim_{\xi \to 0} \xi^{j-k}
        \polylog_{k+1}\left(e^{\frac{-\pi\xi}{2}}\right)
        = \begin{cases} 
            \frac{1}{2} \zeta(k+1) & j = k \\
            0 & j \neq k,
        \end{cases}
\end{split}
\\
\begin{split}
    &\frac{1}{2} \lim_{\xi \to 0} \xi^{j-k}
        \polylog_{k+1}\left(-e^{\frac{-\pi\xi}{2}}\right) \\
        &\quad = \begin{cases} 
            \frac{1}{2} (2^{-k}-1) \zeta(k+1) & j = k \\
            0 & j \neq k,
            \end{cases}
\end{split}
\end{align}
and the relation,
$\zeta(2m) = (-1)^{m+1}(2\pi)^{2m}B_{2m}(2(2m)!)^{-1}$,
between the Riemann Zeta function and the Bernoulli
numbers yields the result of (\ref{eq_xi_csch_solution}):
\begin{align*}
    T_{2m+1}(M) &= \frac{2^{2m+1}(4^{m+1}-1)B_{2m+2}}{(m+1)(2m+1)!} \\
    \quad + &\frac{(-1)^{m+1}4}{\pi} \sum_{k=0}^{2m+1}
        \frac{2^k \tau^{2m+1-k}
        \chi_{k+1}\left(e^{-\frac{\pi \tau}{2}}\right)}
        {\pi^{k}(2m+1-k)!}.
        \label{eq_xi_csch_solution}
\end{align*}

% Can use something like this to put references on a page
% by themselves when using endfloat and the captionsoff option.
\ifCLASSOPTIONcaptionsoff
  \newpage
\fi

%\clearpage

% trigger a \newpage just before the given reference
% number - used to balance the columns on the last page
% adjust value as needed - may need to be readjusted if
% the document is modified later
%\IEEEtriggeratref{73}
% The "triggered" command can be changed if desired:
%\IEEEtriggercmd{\enlargethispage{-3in}}

% references section

% can use a bibliography generated by BibTeX as a .bbl file
% BibTeX documentation can be easily obtained at:
% http://mirror.ctan.org/biblio/bibtex/contrib/doc/
% The IEEEtran BibTeX style support page is at:
% http://www.michaelshell.org/tex/ieeetran/bibtex/
\bibliographystyle{IEEEtran}
% argument is your BibTeX string definitions and bibliography database(s)
\bibliography{bibtex/bib/IEEEabrv.bib,bibtex/bib/IEEEbib.bib}{}
%\balance

%
% <OR> manually copy in the resultant .bbl file
% set second argument of \begin to the number of references
% (used to reserve space for the reference number labels box)

% biography section
% 
% If you have an EPS/PDF photo (graphicx package needed) extra braces are
% needed around the contents of the optional argument to biography to prevent
% the LaTeX parser from getting confused when it sees the complicated
% \includegraphics command within an optional argument. (You could create
% your own custom macro containing the \includegraphics command to make things
% simpler here.)
%\begin{IEEEbiography}[{\includegraphics[width=1in,height=1.25in,clip,keepaspectratio]{mshell}}]{Michael Shell}
% or if you just want to reserve a space for a photo:

%\begin{IEEEbiography}{Mauro J. Sanchirico III}
%Biography text here.
%\end{IEEEbiography}

% if you will not have a photo at all:
%\begin{IEEEbiographynophoto}{Xun Jiao}
%Biography text here.
%\end{IEEEbiographynophoto}

% insert where needed to balance the two columns on the last page with
% biographies
%\newpage

%\begin{IEEEbiographynophoto}{C. Nataraj}
%Biography text here.
%\end{IEEEbiographynophoto}

% You can push biographies down or up by placing
% a \vfill before or after them. The appropriate
% use of \vfill depends on what kind of text is
% on the last page and whether or not the columns
% are being equalized.

%\vfill

% Can be used to pull up biographies so that the bottom of the last one
% is flush with the other column.
%\enlargethispage{-5in}

% that's all folks
\end{document}